\documentclass[sigconf]{acmart}

\usepackage{graphicx}
\usepackage{subcaption}
\usepackage{booktabs}
\usepackage{multicol}
\usepackage{multirow}
\usepackage{hyperref}
\usepackage{url}
\usepackage{graphicx}
\usepackage{algorithm}
\usepackage{algorithmic}
\usepackage{booktabs}
\usepackage{multirow}
\usepackage{color}
\usepackage{wrapfig}
\usepackage{caption}
\usepackage{fontawesome5}
\usepackage{paralist}
\usepackage{mdframed}
\usepackage{soul}

\newenvironment{itemize*}
 {\leftmargini=10pt\begin{compactitem}
  \setlength{\itemsep}{0pt}
  \setlength{\parskip}{0pt}
  }
 {\end{compactitem}}
\newenvironment{enumerate*}
 {\begin{enumerate}
  \setlength{\itemsep}{0pt}
  \setlength{\parskip}{0pt}}
 {\end{enumerate}}

\AtBeginDocument{%
  }

\setcopyright{acmlicensed}
\copyrightyear{2025}
\acmYear{2025}
\acmDOI{XXXXXXX.XXXXXXX}
\acmConference[Conference acronym 'XX]{Make sure to enter the correct
  conference title from your rights confirmation email}{June 03--05,
  2025}{Woodstock, NY}
\acmISBN{978-1-4503-XXXX-X/2025/06}

\begin{document}

\copyrightyear{2025} 
\acmYear{2025} 
\setcopyright{cc}
\setcctype{by}
\acmConference[MM '25]{Proceedings of the 33rd ACM International Conference on Multimedia}{October 27--31, 2025}{Dublin, Ireland}
\acmBooktitle{Proceedings of the 33rd ACM International Conference on Multimedia (MM '25), October 27--31, 2025, Dublin, Ireland}\acmDOI{10.1145/3746027.3754805}
\acmISBN{979-8-4007-2035-2/2025/10}

\title{\textit{MCM-DPO}: Multifaceted Cross-Modal Direct Preference \\Optimization for Alt-text Generation}

\author{Jinlan Fu}
\affiliation{
 \institution{National University of Singapore}
 \city{Singapore}
 \country{Singapore}}
 \email{jinlanjonna@gmail.com}

 \author{Shenzhen Huangfu}
 \authornote{Work done during an internship at NUS.}
\affiliation{
 \institution{Fudan University}
 \city{Shanghai}
 \country{China}}
 \email{shenzhenhuangfu@gmail.com}

 \author{Hao Fei}
\affiliation{
 \institution{National University of Singapore}
 \city{Singapore}
 \country{Singapore}}
 \email{haofei37@nus.edu.sg}

\author{Yichong Huang}
\affiliation{
 \institution{Harbin Institute of Technology}
 \city{Harbin}
 \country{China}}
 \email{ychuang@ir.hit.edu.cn}

 \author{Xiaoyu Shen $^{\dag}$}
\affiliation{
 \institution{Eastern Institute of Technology}
 \city{Ningbo}
 \country{China}}
 \email{xyshen@eitech.edu.cn}

 \author{Xipeng Qiu}
 \authornote{Corresponding author. }
\affiliation{
 \institution{Fudan University}
 \city{Shanghai}
 \country{China}}
 \email{xpqiu@fudan.edu.cn}

\author{See-Kiong Ng}
\affiliation{
 \institution{National University of Singapore}
 \city{Singapore}
 \country{Singapore}}
 \email{seekiong@nus.edu.sg}
 
\renewcommand{\shortauthors}{Jinlan Fu et al.}

\begin{abstract}
  The alt-text generation task produces concise, context-relevant descriptions of images, enabling blind and low-vision users to access online images. Despite the capabilities of large vision-language models, alt-text generation performance remains limited due to noisy user annotations, inconsistent standards, and MLLMs' insensitivity to contextual information.
Previous efforts to fine-tune MLLMs using supervised fine-tuning (SFT) have struggled, as SFT relies on accurate target annotations, which are often flawed in user-generated alt-text. To address this, we propose \ul{Multi-faceted Cross-modal Direct Preference Optimization (MCM-DPO)}, which improves alt-text generation by learning to identify better options in preference pairs without requiring precise annotations.
MCM-DPO optimizes preferences across single, paired, and multi-preference dimensions, covering textual, visual, and cross-modal factors. 
In light of the scarcity of high-quality annotated and preference-labeled datasets for alt-text, we constructed two large-scale, high-quality datasets named TAlt and PAlt, sourced from Twitter and Pinterest. These datasets include 202k annotated alt-text samples and 18k preference pairs that cover diverse preference dimensions, aiming to support further research in this domain. Experimental results show that our proposed MCM-DPO method consistently outperforms both DPO and SFT, establishing a new state of the art in alt-text generation. We release the code and data here:
\url{https://github.com/LVUGAI/MCM-DPO}.
\end{abstract}


\maketitle

\section{Introduction}

Alt-text~\cite{wu2017automatic} refers to a textual description of images, allowing blind or low-vision users to better access visual content, particularly on websites and social media platforms. 
The World Health Organization (WHO) reports that 2.2 billion people worldwide have moderate to severe visual impairments. And according to statistics, up to 98\% images on Twitter still lack alternative text~\cite{srivatsan2023alt}.
The alt-text generation task thus holds significant practical value within multimedia applications.
Recent breakthroughs in multimodal large language models (MLLMs) have spurred exploration of their potential uses. \textit{Can we harness these new MLLM technologies to improve alt‑text and benefit people with visual impairments?}

\begin{figure}
    \centering
    \includegraphics[width=0.98\linewidth]{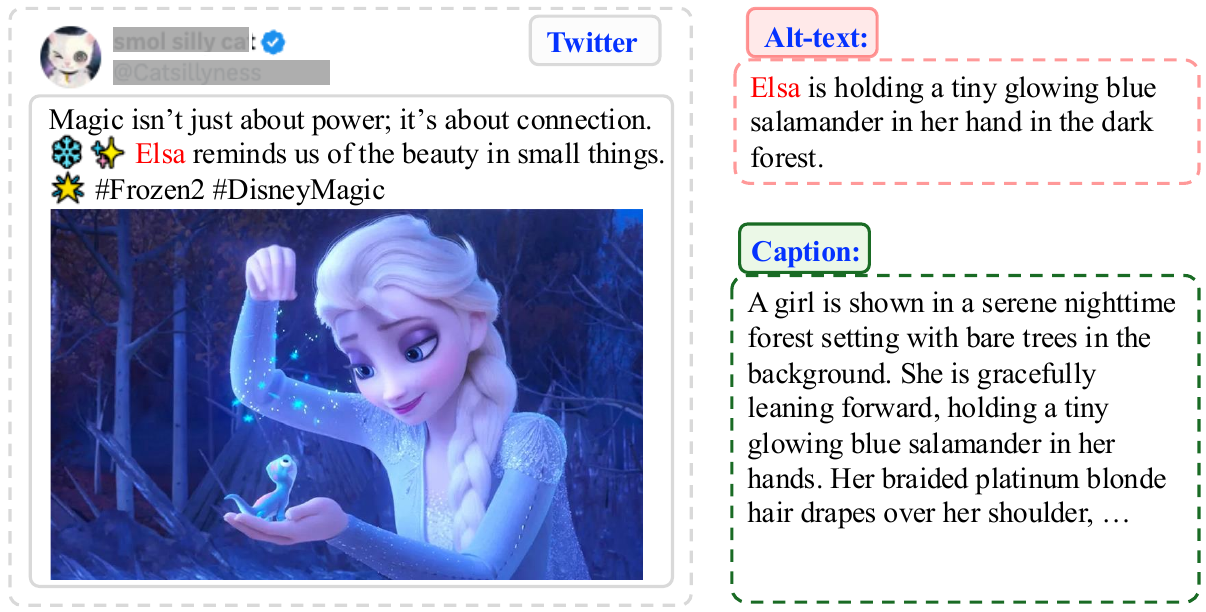}
    \vspace{-8pt}
    \caption{Comparison between alt-text and image caption. In alt-text, ``Elsa'' comes from the post-text (or context). Findings: Alt-text is concise and context-dependent, whereas caption provides detailed descriptions of the image.
    }
    \label{fig:intro}
\end{figure}

There has been extensive exploration of applying MLLMs to image captioning (~\cite{zakir@image-caption}). Whereas image‑captioning models tend to produce detailed, often redundant descriptions, alt‑text generation prioritizes concise, context‑aware summaries that focus on key information (e.g., \autoref{fig:intro}).
Many methods have been explored for alt-text generation, ranging from early sophisticated neural models~\cite{belle2022alt,chintalapati2022dataset,li2020widget,gleason2020twitter} to more recent MLLM-based approaches~\cite{srivatsan2023alt,mohanbabu2024context}.
While MLLMs have achieved state-of-the-art (SoTA) performance, several critical limitations persist across existing methods.
First, most MLLMs are optimized for the image captioning task, which biases them toward producing detailed descriptions instead of the concise, contextually relevant alt-text required~\citep{pmlr-v37-xuc15,Vinyals2014ShowAT}.
Additionally, alt-text data is typically annotated manually by users, introducing considerable noise due to inconsistent labeling criteria among annotators. 
This noisy data can significantly impair MLLM performance.
Finally, these MLLMs rely on supervised fine-tuning (SFT) using large-scale alt-text datasets for training. However, manually annotating alt-text is highly labor-intensive, and the lack of such data limits further performance improvements.

\begin{figure}
  \centering
  \begin{subfigure}{0.45\linewidth}
    \includegraphics[width=1\textwidth]{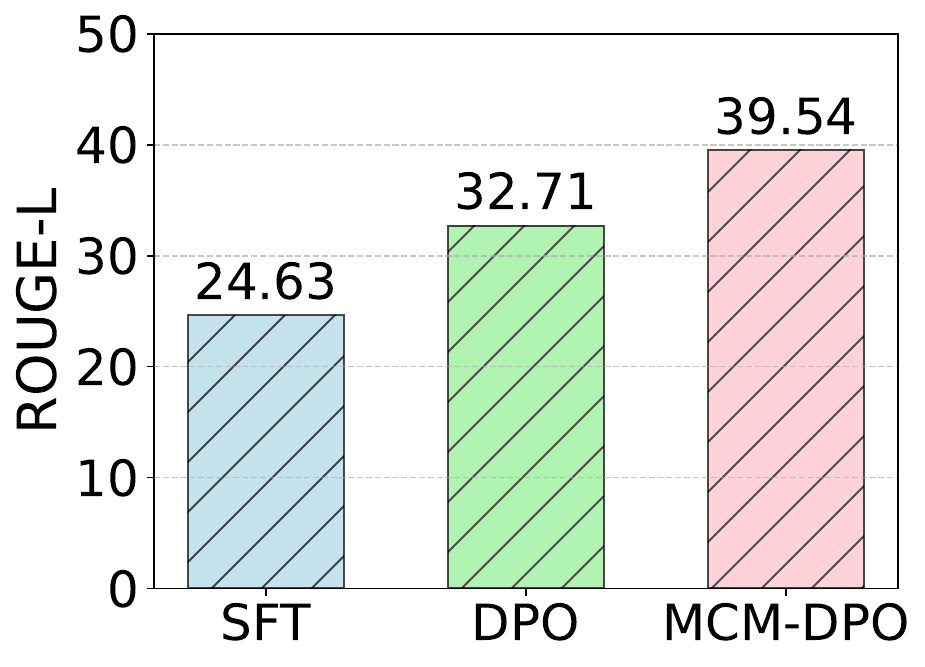}
    \caption{ROUGE-L}
  \end{subfigure}
  \hfill
  \begin{subfigure}{0.45\linewidth}
    \includegraphics[width=1\textwidth]{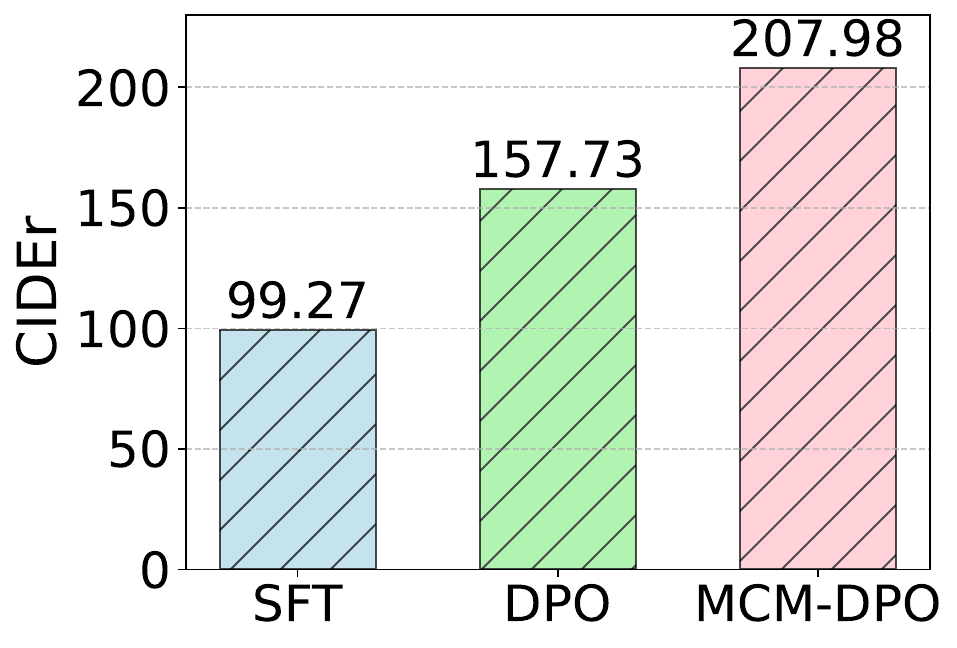}
    \caption{CIDEr}
  \end{subfigure}
  \vspace{-8pt}
  \caption{Results of different models on our PAlt evaluation datasets. SFT: supervised fine-tuned based on LLaVA on 202K human-annotated alt-text samples;  DPO and MCM-DPO: preference optimizations based on the SFT model using the 8K preference dataset.
}
\label{fig:dpo_perfom}
\end{figure}

Recently, Direct Preference Optimization (DPO)~\cite{rafailov2024direct} has emerged as a promising alternative to SFT, aiming to train models to prefer responses that evaluators choose over rejected ones when presented with a user query.
DPO only requires identifying which response better aligns with human preferences, without needing precise target outputs (e.g., alt-text).
Consequently, DPO has proven effective in aligning models with human values~\cite{zhou2023beyond,rafailov2024direct,fuchip,liang2025investigating}, while reducing the reliance on strict annotations—an advantage particularly useful in managing the noisy data prevalent in social media.
This attribute of DPO also helps to mitigate the challenge of huge data reliance.
As shown in \autoref{fig:dpo_perfom}, DPO can significantly mitigate the shortcomings of SFT, improving alt-text generation performance while reducing the amount of training data required.

\begin{figure*}
    \centering
    \includegraphics[width=0.95\linewidth]{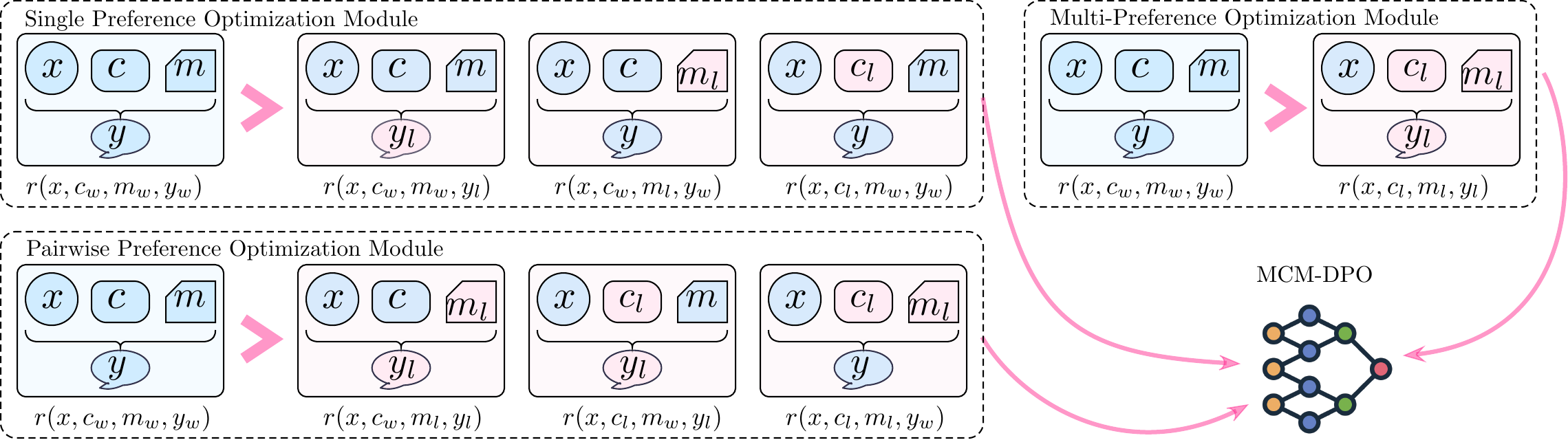}
    \vspace{-8pt}
    \caption{The framework of the MCM-DPO. Symbols $y$, $c$, and $m$ represent alt-text, post-text (context), and image, respectively. The subscript `$_w$' indicates the chosen one (e.g.,$c_w$ denotes the chosen post-text), while `$_l$' denotes the rejected one (e.g., $c_l$ denotes the rejected post-text). By default, y, c, and m refer to $y_w$, $c_w$, and $m_w$.
    }
    \label{fig:framework}
\end{figure*}

However, directly applying DPO to alt-text generation falls short due to the task’s multimodal nature and complex preferences. DPO targets text-only alignment, which overlooks visual content alignment. Moreover, alt-text preferences span single, pairwise, and multi-level aspects—without modeling these granularly, performance remains sub-optimal.
To fill this gap, in this work, we propose a novel Multifaceted Cross-Modal Direct Preference Optimization (\textbf{MCM-DPO}) for alt-text generation.
In our MCM-DPO architecture (as shown in \autoref{fig:framework}), the preference optimization covers Single Preference, Pairwise Preference, and Multi-Preference, spanning seven specific dimensions across text, image, and cross-modal aspects.
With MCM-DPO to build preferences from multiple perspectives, we reach the goal of offering richer and more diverse training signals, helping the model to learn the complex relations between different modalities for better alt-text generation.

Further, to address the lack of multifaceted preference datasets, we propose an alt-text annotation dataset in this work.
We collect alt-text annotations from various users on Twitter \cite{twitter} and Pinterest \cite{pinterest}, and with the assistance of Gemini~\cite{gemini},
we generate high-quality preference and evaluation datasets, named \textbf{TAlt} for Twitter Alt-text and \textbf{PAlt} for Pinterest Alt-text.
Furthermore, we gather a large-scale alt-text dataset consisting of 202K samples to support SFT prior to preference optimization, facilitating adaptation to new domains and tasks.

Experimental results on the TAlt and PAlt datasets demonstrate that our MCM-DPO consistently outperforms both DPO and SFT methods.
In-depth analyses further reveal insights into why our approach works effectively.
Overall, the main contributions of our work are four-fold:
\begin{itemize}
    \item We propose a novel method, MCM-DPO, to reduce the reliance on accurate target alt-text annotations in SFT-based methods. It optimizes preferences across seven aspects based on alt-text, images, post-text (context), and their combinations, offering diverse training signals.
    \item To address the scarcity of large-scale annotated alt-text and preference pairs, we curated two high-quality multimodal datasets, TAlt and PAlt, from Twitter and Pinterest, comprising 202K alt-text samples and 18K preference-annotated samples, supporting future research in this area.
    \item We thoroughly explored four training paradigms to evaluate their effectiveness when MLLMs face new tasks and domains.
    \item Empirically, our MCM-DPO significantly outperforms strong baselines on the TAlt and PAlt evaluation datasets, setting a new SoTA performance for alt-text generation.
\end{itemize}

\section{Related Work}

\paragraph{Alt-text Generation}
The task has attracted extensive research attention in the related area.
Existing methods can be divided into two types: similarity matching~\cite{gleason2020twitter} and abstractive generation~\cite{wu2017automatic,belle2022alt,li2020widget}.
Twitter Ally~\cite{gleason2020twitter} scrapes alt-text from various sources, generating captions only if other methods fail.
Automatic Alt-Text (AAT)~\cite{wu2017automatic} differs by outputting a list of tags for detected objects instead of generating text abstractively.
Alt-Textify~\cite{belle2022alt} introduced a method for generating alt-text for SVG visualizations and charts.
\citet{li2020widget} focused on alt-text for mobile UI elements.

\paragraph{Direct Preference Optimization.}
There has been extensive research related to DPO (Direct Preference Optimization). Existing studies often focus on modifying components of DPO (e.g., reward models or reference models) or improving the quality of training datasets to further enhance the performance of DPO training.
IPO~\cite{IPO} aims to address the reward fitting problem in DPO, while CHiP~\cite{fuchip}, D2PO~\cite{wang2025world}, and PanoDPO~\cite{liang2025investigating} seek to improve preference learning by constructing more preference features.
Methods such as CPO~\cite{CPO}, TPO~\cite{TPO}, and SimPO~\cite{SimPO} attempt to eliminate the use of reference models, simplifying DPO to improve its efficiency. Other methods, such as Iterative DPO~\cite{iter-dpo-yuan}, SSPO~\cite{SSPO}, and WPO~\cite{WPO}, focus on enhancing the training dataset to improve DPO performance. 

In this work, we focus on the issues of inconsistency and high noise in the Alt-text annotations of images on social media, as the annotators are typically the users who post the images. We attempt to construct a preference optimization framework (MCM-DPO) spanning three aspects and nine dimensions. Experimental results demonstrate that our MCM-DPO significantly outperforms large-scale supervised fine-tuning baselines.

\section{Multifaceted Cross-Modal Direct Preference Optimization}
\label{sec:method}
In this work, we propose Multifaceted Cross-modal Direct Preference Optimization (MCM-DPO) to thoroughly explore DPO in the multimodal scenario and develop a tailored method for aligning human preferences for the alt-text generation task.

For the Alt-text generation task, each sample contains a prompt $x$, an image ($m$), context ($c$), and a response $y$. Unlike previous multimodal DPO approaches that only consider preference pairs based on the response (here is alt-text), our MCM-DPO establishes preference pairs based on the image, context, response, and their combination. 
Specifically, the image ($m$) has a chosen ($m_w$) and rejected ($m_l$) image pair, the context ($c$) has a chosen ($c_w$) and rejected ($c_l$) context pair, and the response ($y$) has a chosen ($y_w$) and rejected ($y_l$) response pair.
MCM-DPO involves the design of seven different reward functions, covering three aspects design:
\textit{Single Preference Optimization Module} (\autoref{sec:single}): designs the reward function by considering only one pair of preferences (e.g., $(c_w,c_l)$);
\textit{Pairwise Preference Optimization Module} (\autoref{sec:pairwise}): designs the reward function by considering two out of the three pairs of preferences (e.g., $(c_w,c_l)$ and $(y_w,y_l)$; 
\textit{Multi-Preference Optimization Module} (\autoref{sec:multi}): designs the reward function by considering all three pairs of preferences.
The MCM-DPO framework is shown in \autoref{fig:framework}.
Next, we will provide a detailed introduction to MCM-DPO.

\subsection{Preliminaries}

Direct Preference Optimization (DPO)~\cite{rafailov2024direct} is a technique to optimize a language model directly based on preferences.
Some studies have directly adapted DPO from text modality to multimodal by simply replacing textual preference data with multimodal preference data~\cite{pi@bootstrapped,sarkar@augment}.
Each sample includes an image ($m$) in addition to prompt $x$, chosen response $y_w$, and rejected response $y_l$. DPO relies on both prompt $x$ and image $m$ to select the preferred response from $\{y_w, y_l\}$. Therefore, the DPO in the MLLMs scenario is tried to maximize $\sigma(r(x, m_w, y_w)- r(x, m_w, y_l))$, and the objective function can be denoted as:
\begin{equation}\label{eq:mdpo}
\resizebox{.9\hsize}{!}{%
$\begin{aligned}
    \mathcal{L_{MDPO}} &= -\mathbb{E}\bigl[\log \sigma(r(x, m_w, y_w)- r(x, m_w, y_l))\bigr] \\
    &=  -\log \sigma \bigg(\beta\log \frac{\pi_{\theta} (y_w|m_w,x)}{\pi_\text{ref} (y_w|m_w,x)}- \beta\log \frac{\pi_{\theta} (y_l|m_w,x)}{\pi_\text{ref} (y_l|m_w,x)}\bigg).
\end{aligned}$%
}
\end{equation}

\begin{figure*}
    \centering
    \includegraphics[width=0.9\linewidth]{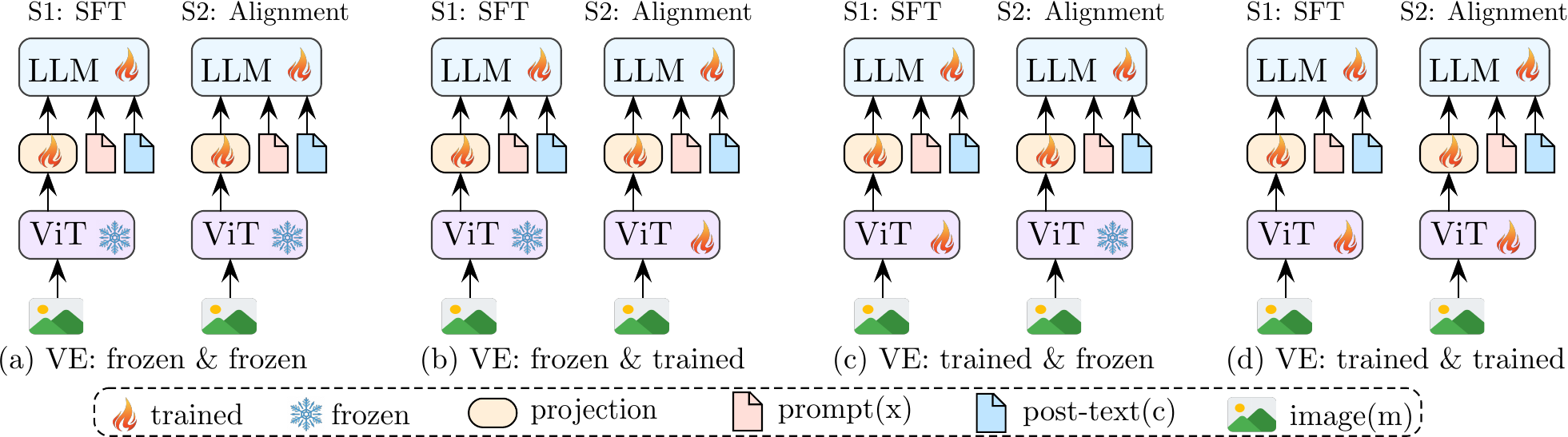}
    \vspace{-8pt}
    \caption{Training paradigms explored in this work. S1 and S2 denote the `Stage 1: supervised fine-tuning' and `Stage 2: preference optimization', respectively.
    }
    \label{fig:training_paradigms}
\end{figure*}

\subsection{Single Preference Optimization Module}
\label{sec:single}
Single Preference Optimization includes \textit{Responsive Preference Optimization}, \textit{Visual Preference Optimization}, and \textit{Contextual Preference Optimization}, each constructing the reward function by establishing a single preference pair.

\vspace{6pt}
\noindent
\textbf{Responsive Preference Optimization}
Our response preference optimization is adapted from multimodal DPO to Alt-text generation, in addition to the prompt $x$, image $m$, and response $y$, also includes context $c$. 
Based on \autoref{eq:mdpo}, the response preference optimization can be denoted as:

\vspace{7pt}
\noindent
\resizebox{.98\hsize}{!}{
$\begin{aligned}
    \mathcal{L_{RPO}} = -\mathbb{E}\bigl[\log \sigma(r(x, m_w, c_w, y_w)- r(x, m_w, c_w, y_l))\bigr].
\end{aligned}$
}
\vspace{7pt}

\noindent
For the chosen response $y_w$, we collected human preferences from Twitter and Pinterest platforms, while for the rejected response $y_l$, we collected it with Gemini's assistance.

\vspace{6pt}
\noindent
\textbf{Visual Preference Optimization}
Since preferences in existing multimodal tasks are typically built based on the response, which overlooks the important role of images in preference alignment and modality alignment. Here, we consider building preferences based on images to mitigate this problem.
Given a preference pair of images $(m_w, m_l)$, the objective function of visual preference optimization can be formulated as:

\vspace{7pt}
\noindent
\resizebox{.98\hsize}{!}{
$\begin{aligned}
    \mathcal{L_{VPO}} =  -\mathbb{E}\bigl[\log \sigma(r(x, m_w, c_w, y_w)- r(x, m_l, c_w, y_w))\bigr].
\end{aligned}$
}
\vspace{7pt}

\noindent
$m_l$ can be a transformation of $m_w$, such as rotation.
We explored different strategies for $m_l$ (i.e., rotation) in section~\autoref{sec:reject_image_strategy}.

\vspace{6pt}
\noindent
\textbf{Contextual Preference Optimization}
In the Alt-text generation task, the model generates image descriptions relying on both the image and the context, ensuring the descriptions do not simply repeat the context. Therefore, context plays a crucial role in this task, and we create preference pairs based on the context. Increasing the number of preference optimization targets may help improve the alignment between the image and text modalities.

Given a preference pair of contexts $(c_w, c_l)$, 
the objective function of contextual preference optimization can be formulated as:

\vspace{6pt}
\noindent
\resizebox{.98\hsize}{!}{
$\begin{aligned}
    \mathcal{L_{CPO}} =  -\mathbb{E}\bigl[\log \sigma(r(x, m_w, c_w, y_w)- r(x, m_w, c_l, y_w))\bigr].
\end{aligned}$
}
\vspace{6pt}

\noindent
Here, we construct the $c_l$ by randomly selecting a context from other samples in the training set.

\subsection{Pairwise Preference Optimization Module}
\label{sec:pairwise}
In the single preference optimization module, we aim to construct preference pairs based on a single dimension—response, image, or context. Additionally, we can extend this by selecting any two dimensions, such as response and image, to form preference pairs. This approach enhances the diversity of preferences, promoting more effective alignment between the text and image modalities.

\vspace{6pt}
\noindent
\textbf{Visual and Responsive Preference Optimization}
\label{sec:vrpo}
For visual and responsive preference optimization, we aim to construct preferences based on the image and response dimensions. 
Given a preference pair of images $(m_w, m_l)$ and responses $(y_w, y_l)$, the objective function can be formulated as:

\vspace{7pt}
\noindent
\resizebox{.98\hsize}{!}{
$\begin{aligned}
    \mathcal{L_{VRPO}} =  -\mathbb{E}\bigl[\log \sigma(r(x, m_w, c_w, y_w)- r(x, m_l, c_w, y_l))\bigr].
\end{aligned}$
}

\vspace{6pt}
\noindent
\textbf{Contextual and Responsive Preference Optimization}
For contextual and responsive preference optimization, we focus on constructing preferences along the context and response dimensions. 
Given a pair of contexts $(c_w, c_l)$ and responses $(y_w, y_l)$, the objective function is:

\vspace{7pt}
\noindent
\resizebox{.98\hsize}{!}{
$
\begin{aligned}
    \mathcal{L_{CRPO}} =  -\mathbb{E}\bigl[\log \sigma(r(x, m_w, c_w, y_w) - r(x, m_w, c_l, y_l))\bigr].
\end{aligned}$
}

\vspace{6pt}
\noindent
\textbf{Visual and Contextual Preference Optimization}
The goal of the visual and contextual preference optimization is to establish preferences based on the image and context dimensions. Considering a preference pair of contexts $(c_w, c_l)$ and images $(m_w, m_l)$, the objective function is then defined as:

\vspace{7pt}
\noindent
\resizebox{.98\hsize}{!}{
$\begin{aligned}
    \mathcal{L_{VCPO}} =  -\mathbb{E}\bigl[\log \sigma(r(x, m_w, c_w, y_w) - r(x, m_l, c_l, y_w))\bigr].
\end{aligned}$
}
\vspace{7pt}

\subsection{Multi-Preference Optimization Module}
\label{sec:multi}
This module aimed at establishing preferences across the three dimensions of image, context, and response.
Consider a preference pair of responses $(y_w, y_l)$, images $(m_w, m_l)$, and contexts $(c_w, c_l)$, the objective function can be formulated as:

\vspace{7pt}
\noindent
\resizebox{.98\hsize}{!}{
$\begin{aligned}
    \mathcal{L_{MTPO}} =  -\mathbb{E}\bigl[\log \sigma(r(x, m_w, c_w, y_w) - r(x, m_l, c_l, y_l))\bigr].
\end{aligned}$
}

\subsection{Putting All Modules Together}
By combining the preference optimization objectives from each module, we derive the overall optimization goal for MCM-DPO. This integration of the three preference optimization modules results in a total of seven reward losses. The optimization objective for MCM-DPO is formulated as follows:

\vspace{7pt}
\noindent
\begin{equation}\label{eq:target}
\begin{aligned}
    \mathcal{L_{MCM-DPO}} = \lambda * \mathcal{L_{RPO}} + \alpha * (\mathcal{L_{VPO}} +\mathcal{L_{CPO}}) \\
   \hspace*{-0.1em} + \gamma * (\mathcal{L_{VRPO}} + \mathcal{L_{CRPO}} +\mathcal{L_{VCPO}} + \mathcal{L_{MTPO}}).
\end{aligned}
\end{equation}
\vspace{7pt}

\noindent
Here, $\lambda$, $\alpha$, and $\gamma$ is the weight for different modules.
Through a series of experiments, we found that the best performance is achieved when $\lambda =1$, $\alpha = 0.5$, and $\gamma = 0.2$.

\subsection{Training Paradigms}
\label{sec:paradigms}
MLLMs consist of a vision encoder, LLMs, and a projection.
To reduce computational costs and preserve robust visual features, many works~\cite{yu2024rlhf,wang2024mdpo} typically freeze the vision encoder, focusing training solely on the LLMs and the projection. This restricts the integration of cross-modal information, resulting in hallucinations and other undesirable error outputs.

To explore the impact of training paradigms on MLLMs' performance in new domains (e.g., social media) and tasks, we examine four paradigms, as shown in \autoref{fig:training_paradigms}.
Specifically, our training process is divided into two stages, Stage 1:  supervised fine-tuning (SFT) on large-scale datasets, 
and Stage 2: preference optimization (i.e., DPO or MCM-DPO) on small high-quality human preference datasets. 
The training paradigms explored are listed below:
\begin{itemize}
    \item \textbf{Paradigm-1}: The vision encoder is frozen during both Stage 1 and Stage 2 (\autoref{fig:training_paradigms}-(a));
     \item \textbf{Paradigm-2}: The vision encoder is frozen during Stage 1 but trained during Stage 2 (\autoref{fig:training_paradigms}-(b));
      \item \textbf{Paradigm-3}: The vision encoder is trained during Stage 1 but frozen during Stage 2 (\autoref{fig:training_paradigms}-(c));
       \item \textbf{Paradigm-4}: The vision encoder is trained during both Stage 1 and Stage 2 (\autoref{fig:training_paradigms}-(d)).
\end{itemize}

\begin{figure}
    \centering
    \includegraphics[width=0.96\linewidth]{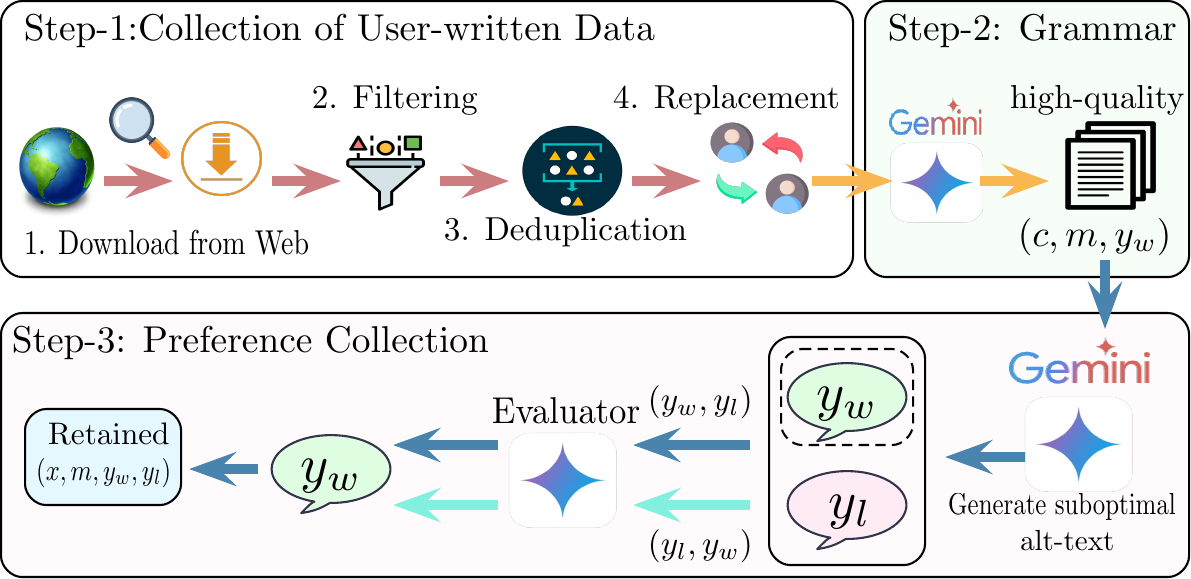}
    \vspace{-8pt}
    \caption{Dataset construction process. $c$ and $m$ represent the post-text (context) and image, respectively.
$y_w$ and $y_l$ are the chosen and rejected alt-text, respectively. 
    }
    \label{fig:data-construction}
\end{figure}

\section{Dataset Construction}
\label{sec:data-construction}
To collect a high-quality dataset of human preferences for building an automatic alt-text generation system, we gathered human-annotated alt-text datasets from two widely used social media platforms: Twitter and Pinterest.
As shown in Fig. \ref{fig:data-construction}, our dataset construction process involves three steps: (1) Collection of User-written Data; (2) Grammar Correction; (3) Preference Collection.

\subsection{Overview}
We collected three datasets on each platform (Twitter and Pinterest), respectively.
(1) \textbf{Large-scale supervised dataset: } contains a total of 202K triplet \textit{(context, image, alt-text)}, which is utilized to initialize a base model with SFT, enabling it to acquire knowledge in social media. It was collected after step-2 (grammatical revisions).
(2) \textbf{Preference optimization dataset: } was constructed with the help of Gemini. It will be used for preference alignment based on the SFT model.
(3) \textbf{Testing dataset: } is the evaluation dataset for the alt-text generation task, named TAlt-Test and PAlt-Test, contains 1.7K samples each from Twitter and Pinterest.
Tab.~\ref{tab:statistic} shows the statistics of our constructed dataset.

\begin{table}[htbp]
  \centering  
      \caption{Statistics of the datasets constructed on Twitter (TAlt) and Pinterest (PAlt), including three parts: large-scale SFT set, preference optimization dataset, and test set.}
  \label{tab:statistic}
  \vspace{-8pt}
    \begin{tabular}{lccc}
    \toprule
    Dataset & Training & Preference & Test \\
    \midrule
    PAlt   & 101,715 & 8,212 & 1,702 \\
    TAlt  & 101,078 & 9,938 & 1,688 \\
    \bottomrule
    \end{tabular}
\end{table}

\subsection{Step-1: Collection of User-written Data}

\textbf{Collection from Web}
\texttt{Pinterest:} We scraped context-image pairs with alt-text from Pinterest to collect the user-written alt-text.
To ensure diversity, we used 100,000 commonly used words from Wikipedia as query keywords to find the web pages. 
According to statistics~\cite{srivatsan2023alt}, only about 2\% of images on social media platforms (e.g., Twitter) have manually annotated alt-text. Since Pinterest does not provide an API, we utilized common words to search and collect over 22 million posts from Pinterest, filtering approximately 150,000 rough samples containing images, post-text, and alt-text. \ul{This indicates that only 0.68\% of samples on Pinterest include alt-text, underscoring the difficulty of collecting manually annotated alt-text data. Therefore, we leave the investigation of potential biases in the collected dataset, such as cultural background, and gender, for future research.}
\texttt{Twitter:}
We leveraged the publicly available Twitter alt-text dataset from \cite{srivatsan2023alt}, which was collected using the Twitter API, contains 371K pairs of alt-text, images, and tweets.

\noindent
\textbf{Filtering}
To ensure the initial quality of the data, we first filter out low-quality samples.
We filtered out non-static images (e.g., GIFs), samples with non-English, and hashtags. Additionally, we restricted the alt-text length to be at least 5 words.

\noindent
\textbf{Deduplication}
To ensure the diversity of the collected samples, we used CLIP to filter out duplicate data based on the representation similarity of post-text or image embeddings across different samples.
Specifically, we determined visual match clusters based on pixel overlap. 

\noindent
\textbf{Person Name Replacement}
Predicting a person's name from an image involves techniques such as facial recognition, which is beyond the scope of this study. 
We use the named entity recognition tools to identify named entities and replacing them with ``[person]''.

\subsection{Step-2: Grammar Correction}
User-generated text is often noisy and irregular. To reduce noise introduced by users, we performed grammar correction on the textual components of the collected samples, including post-text and alt-text.
Specifically, we used \textit{geimini-1.5-flash}\cite{gemini} as the grammar corrector.
The grammar correction prompt is shown in the appendix.

\begin{table*}[!t]
  \centering 
  \caption{Results of alt-text generation in the Twitter (TAlt) and Pinterest (PAlt) domains. The values in \textbf{bold} denote the best performance among SFT, DPO, and MCM-DPO.
       ``+SFT'' denotes the model is supervised fine-tuning on our 202k training set.
      }
  \label{tab:alt-res}
  \vspace{-8pt}
    \begin{tabular}{lcccccccc}
    \toprule
    \multirow{1}[1]{*}{Model}  & \multicolumn{4}{c}{PAlt (Pinterest)} & \multicolumn{4}{c}{TAlt (Twitter)} \\
  \cmidrule(r){2-5} \cmidrule(l){6-9} 
              & ROUGE-L & BLEU4 & METEOR & CIDEr & ROUGE-L & BLEU4 & METEOR & CIDEr \\
    \midrule
    InstructBLIP (13B)~\cite{InstructBLIP}      & 5.99  & 0.05  & 7.17  & 1.90  & 7.67  & 0.08  & 8.25  & 1.95 \\
    Chameleon ~\cite{team2024chameleon}      & 13.69 & 0.74  & 24.02 & 21.58 & 12.51 & 0.37  & 14.27 & 10.49 \\
    Qwen-VL-Chat (9.6B)~\cite{Qwen-VL}      & 19.03 & 2.72  & 35.11 & 50.11 & 18.29 & 3.26  & 20.60 & 37.49 \\
    MiniCPM-V 2.0 (2.8B)~\cite{hu2024minicpm}      & 20.15 & 3.37  & 32.28 & 55.16 & 21.29 & 3.28  & 22.53 & 41.25 \\
    MiniCPM-Llama3-V 2.5 (8B)~\cite{yao2024minicpmv}      & 18.04 & 1.44  & 33.36 & 34.40 & 17.67 & 1.56  & 20.95 & 25.18 \\
    LLaVA-NEXT-Vicuna (13B)~\cite{liu2023improved}      & 19.48 & 3.29  & 33.18 & 51.55 & 21.30 & 4.94  & 22.39 & 51.13 \\
    LLaVA-NEXT (34B)~\cite{liu2023improved}      & 20.99 & 2.76  & 31.23 & 57.97 & 19.41 & 3.44  & 19.84 & 42.20 \\
    VIP-LLaVA (13B)~\cite{cai2023making}      & 20.50 & 3.27  & 33.49 & 53.87 & 20.47 & 3.10  & 20.18 & 41.37 \\
    LLaVA-1.6 (7B)     & 17.32 & 2.46  & 31.46 & 40.23 & 20.71 & 4.86  & 21.31 & 49.85 \\
    \cmidrule(lr){1-9}
    LLaVA-1.6 (7B)+SFT     & 24.63 & 7.63  & \textbf{41.13} & 99.27 & 33.83 & 14.90 & 33.29 & 144.64 \\
    \quad+DPO        & 32.71 & 11.11 & 40.84 & 157.73 & 35.32 & 15.87 & 36.32 & 146.32 \\
    \quad+mDPO~\cite{wang2024mdpo}  & 36.63 & 12.90 & \textbf{41.37} & 183.89 & 35.74 & 15.78 & 35.77 & 153.15 \\
   \quad+MCM-DPO       & \textbf{39.54} & \textbf{14.48} & 40.60 & \textbf{207.98} & \textbf{36.32} & \textbf{16.25} & \textbf{36.62} & \textbf{158.74} \\
    \bottomrule
    \end{tabular}
\end{table*}

\subsection{Step-3: Preference Collection}

To construct preference data, we need to create the suboptimal alt-text for each image to obtain the preference pair.

\noindent
\textbf{Rejected Alt-text Generation}
To construct preference data, we used Gemini (gemini-1.5-flash) to generate suboptimal alt-texts, forming pairs where the chosen response is human-preferred. \ul{To reduce bias from using a single model, we applied temperature sampling and manual filtering} to ensure rejected responses are clearly inferior in accuracy and coherence, without enforcing stylistic patterns. As preference learning relies on relative ranking, model-specific bias has limited effect.
The prompt for the generation can be found in the appendix.

\noindent
\textbf{Quality Verification}
To verify that $x_w$ and $x_l$ have a clear preference, we use Gemini as the evaluator. To eliminate position bias, we swap the positions of $x_w$ and $x_l$. If Gemini considers $x_w$ to be the chosen alt-text both before and after swapping positions, the ($x_w$, $x_l$) pair will be retained. 
Otherwise, Gemini will regenerate $x_l$ and re-evaluate the quality by swapping the positions of $x_l$ and $x_w$ until it consistently selects $x_w$ as the alt-text after the swap (with a maximum of three attempts).

\section{Experiment and Results}

\subsection{Experimental Settings}

\noindent
\subsubsection{Baselines}
We primarily compare MCM-DPO with standard DPO using the same underlying models.
We considered the following open MLLMs as our baselines:
InstructBLIP (Vicuna-13B)~\cite{InstructBLIP}, 
Qwen-VL-Chat~\cite{Qwen-VL}, 
MiniCPM-V 2.0~\cite{hu2024minicpm}, 
MiniCPM-Llama3-V 2.5~\cite{yao2024minicpmv}, 
LLaVA-NEXT (Vicuna, 34B) ~\cite{liu2023improved},  
VIP (LLaVA, 13B)~\cite{cai2023making}, and 
Chameleon (30B)~\cite{team2024chameleon}.
To guarantee reproducibility, we eliminate randomness 
by setting the temperature to 0.
To better match the alt-text task, we guide the baseline models to generate shorter outputs by adding two example alt-texts to the inference prompt.

\noindent
\subsubsection{Metrics}
Like image captioning task~\cite{zakir@image-caption}, we employ automatic evaluation metrics.
Specifically, we use user-written alt-text as a reference and compute string similarity with the model-generated text.
The metrics we consider include 
ROUGE-L~\cite{lin2004automatic}, BLEU4~\cite{papineni2002bleu}, METEOR~\cite{denkowski2014meteor}, and CIDEr~\cite{vedantam2015cider}.

\noindent
\subsubsection{Implementation Details}

We used LLaVA-1.6~\cite{liu2023improved} as the base model, which combines CLIP's ViT~\cite{radford2021learning} as vision encoder and Vicuna-7B~\cite{liu2023improved} as language backbone. The model was supervised fine-tuning (SFT) for one epoch on 202K samples from Twitter (TAlt) and Pinterest (PAlt) using a resolution of 336, learning rate 2e-5, and batch size 128, enabling effective domain adaptation. See appendix for details.

\noindent
\textit{Preference optimization}
We further applied DPO and MCM-DPO on the SFT model using PAlt and TAlt preference datasets, training for three epochs with a 5e-7 learning rate and batch size 64.

\noindent
\textit{License}
All data, code, and models are for research use only, following LLaMA and Vicuna license terms.

\subsection{Main Results and Analysis}

The results of our MCM-DPO based on LLaVA-1.6 and baseline models on the PAlt and TAlt datasets are shown in \autoref{tab:alt-res}.
The main findings are summarized as below:

\noindent
(1) \textit{Baseline models that excel in generating general image descriptions perform poorly on the alt-text generation task.} 
This indicates that SOTA multimodal models are not well-suited for alt-text generation, which requires context dependency.

\noindent
(2) \textit{LLaVA exhibits significant improvements following SFT on large-scale domain-specific data.} Despite heavy user noise, PAlt and TAlt still help the model adapt to the new domain and task.

\noindent
(3) \textit{MCM-DPO outperforms both SFT and DPO significantly.} 
This suggests MCM-DPO's success stems from its multi-perspective cross-modal preference optimization.

\begin{table*}[htb]
  \centering 
  \caption{ Performance under different training paradigms.
The values in \textbf{bold} denote the best performance. ``+SFT'' denotes the model is supervised fine-tuning on our 202k training set. $\checkmark$ indicates that the parameters of the visual encoder are trained in the first or second stage, $\times$ indicates that the visual encoder is frozen, and - indicates not applicable.
      }
  \label{tab:train_paradigm}
  \vspace{-8pt}
    \begin{tabular}{lcccccccccc}
    \toprule
    \multirow{2}[1]{*}{Model} & \multicolumn{2}{c}{Visual Encoder} & \multicolumn{4}{c}{PAlt (Pinterest)} & \multicolumn{4}{c}{TAlt (Twitter)} \\
    \cmidrule(lr){2-3} \cmidrule(lr){4-7} \cmidrule(lr){8-11} 
          & SFT (S1) & Align (S2) & ROUGE-L & BLEU4 & METEOR & CIDEr & ROUGE-L & BLEU4 & METEOR & CIDEr \\
          \midrule
           LLaVA-1.6 (7B) & -     & - & 17.32 & 2.46  & 31.46 & 40.23 & 20.71 & 4.86  & 21.31 & 49.85 \\
           \cmidrule(lr){1-11}
    +SFT & $\times$     & -     & 24.73 & 7.41  & \textbf{40.85} & 97.63 & 34.45 & 14.98 & 33.91 & 143.83 \\
    \quad+DPO & $\times$     & $\times$     & 32.30 & 10.86 & 40.85 & 153.10 & 35.15 & 15.49 & 36.39 & 151.42 \\
    \quad+MCM-DPO & $\times$     & $\times$     & 36.73 & \textbf{12.56} & 37.37 & \textbf{186.97} & 35.20 & 15.45 & 35.59 & 151.61 \\
    \quad+DPO   & $\times$     & $\checkmark$     & 31.33 & 10.50 & 41.00 & 147.05 & 35.22 & 15.50 & \textbf{36.63} & 150.97 \\
    \quad+MCM-DPO & $\times$     & $\checkmark$     & 36.51 & 12.34 & 37.07 & 184.80 & \textbf{35.44} & \textbf{15.67} & 35.79 & \textbf{153.10} \\
    \midrule
    +SFT  & $\checkmark$     & -     & 24.63 & 7.63  & \textbf{41.13} & 99.27 & 33.83 & 14.90 & 33.29 & 144.64 \\
    \quad+DPO  & $\checkmark$     & $\times$     & 32.27 & 11.07 & 40.90 & 155.33 & 35.18 & 15.67 & 36.13 & 144.68 \\
    \quad+MCM-DPO & $\checkmark$     & $\times$     & 39.29 & 14.32 & 39.87 & 206.01 & 35.96 & 16.06 & 36.34 & 157.21 \\
    \quad+DPO  & $\checkmark$     & $\checkmark$     & 32.71 & 11.11 & 40.84 & 157.73 & 35.32 & 15.87 & 36.32 & 146.32 \\
    \quad+MCM-DPO & $\checkmark$     & $\checkmark$     & \textbf{39.54} & \textbf{14.48} & 40.60 & \textbf{207.98} & \textbf{36.32} & \textbf{16.25} & \textbf{36.62} & \textbf{158.74} \\
    \bottomrule
    \end{tabular}
\end{table*}

\subsection{Impact of Training Paradigm}
We thoroughly investigate the impact of different training paradigms (as discussed in \autoref{sec:paradigms}) on the effectiveness of preference optimization, and the results are shown in \autoref{tab:train_paradigm}. Observations:

\noindent
(1) \textit{Keeping the vision encoder trainable during both the SFT and MCM-DPO stages yields the best performance. }
The performance difference between training and freezing the vision encoder during SFT is minimal (ROUGE-L on Pinterest: 24.73 vs. 24.63), but more significant improvements are seen after MCM-DPO.
The possible reason is that updating the vision encoder during the SFT stage likely activates neurons that improve task-specific feature extraction, which then contribute to preference optimization.

\noindent
(2) \textit{The effectiveness of training paradigms is highly influenced by the preference optimization method used.}
For DPO, making the vision encoder trainable during SFT has little impact, but for MCM-DPO, it significantly improves results when based on SFT (VE-trained).  This may be due to MCM-DPO's broader design, which considers seven preference aspects, unlike DPO's single focus.

\subsection{Human Evaluation}
\begin{figure}[ht!]
    \centering
    \includegraphics[width=0.7\linewidth]{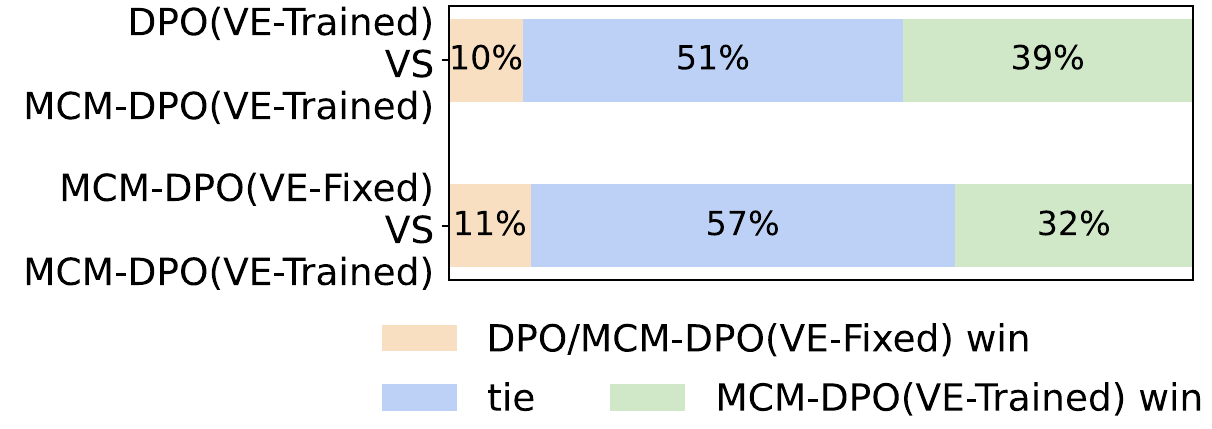}
    \vspace{-8pt}
    \caption{Human Evaluation.
    }
    \label{fig:human}
\end{figure}

To improve evaluation reliability, we also conducted human evaluation, as shown in \autoref{fig:human}. 
Given the cost and time of human evaluation, we focused on:
(1) \textit{The effectiveness of MCM-DPO versus DPO}, using paradigm-4 with the vision encoder trainable in both stages.
(2) \textit{Training paradigm effectiveness}, comparing MCM-DPO in paradigm-4 (vision encoder trainable in both stages) to paradigm-1 (vision encoder frozen in both stages).

\noindent
\textbf{Findings:}
We found that MCM-DPO outperformed DPO by 29\% (39\% vs. 10\%) and that training paradigm 4 exceeded paradigm 1 by 21\% (32\% vs. 11\%). This indicates that keeping the vision encoder trainable during both the SFT and preference alignment stages is an effective approach for training MLLMs on new tasks and domains, leading to notable improvements.

\subsection{Effect in Hallucination Mitigation Task}

To evaluate the effectiveness of the proposed MCM-DPO framework on other public benchmarks, we adapt it for multimodal hallucination tasks using the LLaVA-1.6 backbone and train on RLHF-V~\cite{yu2024rlhf}. Evaluations are on ObjHal~\cite{rohrbach2018object}, MMHal~\cite{sun2023aligning}, and AMBER~\cite{wang2023llm}. Training details are in the appendix.

Results are shown in Table \ref{tab:hallu}. \textbf{Findings:}
\textbf{MCM‑DPO slashes hallucinations without sacrificing quality}: on AMBER the hallucination rate falls to 27.9\% (‑42.6\% relative) while object coverage holds steady, with similar gains on MMHal and the lowest response‑ and mention‑level rates on ObjHal. This shows the method excels on alt‑text and other public benchmarks alike.

\begin{table}[htbp]
  \centering \small
  \setlength{\tabcolsep}{2pt}
  \caption{Results of MCM-DPO on multimodal hallucination. Bold is the best performance. `↓' means lower is better, while `↑'means higher is better. \texttt{M.}, \texttt{R.}, and \texttt{O.} denote the mention-level hallucination rate, response-level hallucination rate, and overall performance, respectively.}
  \vspace{-8pt}
    \begin{tabular}{lcccccccc}
    \toprule
    \multirow{2}[2]{*}{Model} & \multicolumn{2}{c}{ObjHal} & \multicolumn{2}{c}{MMHal} & \multicolumn{4}{c}{AMBER} \\
    \cmidrule(lr){2-3}\cmidrule(lr){4-5}\cmidrule(lr){6-9}
          & R.↓ & M.↓ & O.↑ & R.↓ & CHAIR↓ & Cover↑ & Hal↓  & Cog↓ \\
    \midrule
    GPT-4V~\cite{gpt4V} & 13.6 & 7.3 & - & 31.3 & 4.6 & 67.1 & 30.7 & 2.6 \\
    \midrule
    LLaVA-1.6  & 14.1  & 7.4   & 2.8   & 42.7  & 8.3   & 61.0 & 48.6  & 4.2 \\
    +DPO  & 11.0  & 6.6   & 2.7   & 43.8  & 5.9   & 61.0 & 38.9  & 3.0 \\
    +MCM-DPO & \textbf{8.9}	     & \textbf{5.2 }    & \textbf{2.9}	     & \textbf{38.5}     & \textbf{4.7}     & \textbf{61.2}     & \textbf{27.9}     & \textbf{2.3} \\
    \bottomrule
    \end{tabular}
  \label{tab:hallu}
\end{table}

\section{Further Analyses and Discussions}

\subsection{Rejected Image Construction Strategy}
\label{sec:reject_image_strategy}
\textbf{Strategy}
To enable cross-modal alignment with images as the sole visual modality, we explore five strategies for constructing rejected images based on the chosen image:
(1) \textit{Diffusion \cite{diffusion_noise}}:  add Gaussian noise to the chosen image for T steps;
(2) \textit{Blackness}: set all RGB values to 0;
(3) \textit{Crop}: randomly crop the chosen image;
(4) \textit{Rotation}: rotate the image by 10–80 degrees;
(5) \textit{Randomness}: select a different random image from the training set.

\noindent
\textbf{Results}
The experimental results are presented in \autoref{tab:reject_strategy}.
Our main observation is: \textit{Greater similarity between the rejection and chosen images generally enhances preference optimization with MCM-DPO.}

\noindent
\textbf{Impact of Noise Step T}
We then examined how different noise steps  $T$  affect MCM-DPO (see \autoref{tab:noise_step}) and found performance peaks at $T=700$. Fewer steps make images too similar; too many make them too distinct, limiting visual learning.

\begin{table}[h]
  \centering \small
      \caption{
      MCM-DPO results on the PAlt dataset using different rejected image strategies.  \textbf{Bold} indicates the best performance; The noise step $T=700$ for the \texttt{Diffusion} strategy.
  }
  \label{tab:reject_strategy}
  \vspace{-8pt}
    \begin{tabular}{lcccc}
    \toprule
    Strategy & ROUGE-L & BLEU4 & METEOR & CIDEr \\
    \midrule
    Diffusion & \textbf{39.54} & \textbf{14.48} & \textbf{40.60} & \textbf{207.98} \\
    Blackness & 39.41 & 14.29 & 40.41 & 200.39 \\
    Crop & 39.15 & 14.18 & 39.02 & 205.17 \\
    Rotation & 37.23 & 12.78 & 37.47 & 191.37 \\
    Randomness & 37.63 & 13.51 & 39.44 & 194.71 \\
    \bottomrule
    \end{tabular}
\end{table}

\begin{table}[!h]
  \centering  \small
\caption{Results of MCM-DPO under the \texttt{Diffusion} strategy with different noise step T on the PAlt test set. Values in \textbf{bold} denote achieving the best performance under a metric.}
  \label{tab:noise_step}
  \vspace{-8pt}
  \fontsize{9}{10}\selectfont 
  \setlength{\tabcolsep}{1.7mm}
    \begin{tabular}{ccccc}
    \toprule
    T     & ROUGE-L & BLEU4 & METEOR & CIDEr \\
    \midrule
    300   & 37.00 & 12.52 & 37.40 & 188.29 \\
    500   & 37.53 & 13.15 & 38.00 & 194.55 \\
    700   & \textbf{39.54} & \textbf{14.48} & 40.60 & \textbf{207.98} \\
    800   & 39.06 & 14.31 & 40.97 & 205.37 \\
    1000  & 39.45 & 14.03 & \textbf{41.35} & 204.17 \\
    \bottomrule
    \end{tabular}
\end{table}

\subsection{Effect of Component Combination}
We examined the effectiveness of different MCM-DPO components, evaluating on the PAlt dataset shown in \autoref{tab:effect_component}.
The main findings are shown below:
(1) Constructing preference pairs that simultaneously consider image, context, and alt-text (i.e., $\mathcal{L_{\text{Multi}}}$) makes a significant contribution to the model's preference optimization. We observe a sharp performance drop when MCM-DPO without the $\mathcal{L_{\text{Multi}}}$ component, with ROUGE-L (CIDEr) decreasing from 39.54 (207.98) to 33.60 (165.18). 
(2) Contextual preference is crucial. Removing it hurts the overall performance a lot.
This is likely because context is a unique element in the alt-text generation task.

\begin{table}[!h]
  \centering \small
   \caption{Results of ablation models on the PAlt dataset. \textbf{Bold} indicates the best performance. $\mathcal{L_{\text{Single}}}$ = $\mathcal{L_{\text{RPO}}} + \mathcal{L_{\text{VPO}}} + \mathcal{L_{\text{CPO}}}$.
  $\mathcal{L_{\text{Pair}}}$ = $\mathcal{L_{\text{VRPO}}} + \mathcal{L_{\text{CRPO}}} + \mathcal{L_{\text{VCPO}}}$. 
  $\mathcal{L_{\text{Multi}}}$ = $\mathcal{L_{\text{MTPO}}}$.
  }
  \label{tab:effect_component}%
  \vspace{-8pt}
    \begin{tabular}{lcccc}
    \toprule
    Model & ROUGE-L & BLEU4 & METEOR & CIDEr \\
    \midrule
    MCM-DPO & \textbf{39.54} & \textbf{14.48} & \textbf{40.60} & \textbf{207.98} \\
    -$\mathcal{L_{\text{Single}}}$ & 37.78 & 13.27 & 38.14 & 195.64 \\
    -$\mathcal{L_{\text{Pair}}}$ & 37.63 & 13.51 & 39.44 & 194.71 \\
    -$\mathcal{L_{\text{Multi}}}$ & 33.60 & 11.65 & 40.33 & 165.18 \\
    -$\mathcal{L_{\text{Single+Pair}}}$   & 37.46 & 13.12 & 37.87 & 193.67 \\
    -$\mathcal{L_{\text{Pair}}}$-$\mathcal{L_{\text{CPO}}}$ & 33.60 & 11.01 & 36.77 & 163.52 \\
    \bottomrule
    \end{tabular}
\end{table}

\begin{table}[!t]
  \centering \small
    \caption{Hyperparameter Exploration. Evaluation results of MCM-DPO on PAlt with different values of $\gamma$. Bold values indicate the best performance. Finding: when $\gamma = 0.2$, MCM-DPO achieved the best performance across all four metrics.}
  \label{tab:parameters}%
  \vspace{-8pt}
    \begin{tabular}{ccccc}
    \toprule
    $\gamma$  & ROUGE-L & BLEU4 & METEOR & CIDEr \\
    \midrule
    0.1   & 39.24 & 14.36 & 39.91 & 205.72 \\
    0.2   & \textbf{39.54} & \textbf{14.48} & 40.60 & \textbf{207.98} \\
    0.4   & 37.29 & 13.33 & 40.84 & 192.43 \\
    0.6   & 36.19 & 12.63 & \textbf{40.95} & 182.00 \\
    0.8   & 35.69 & 12.36 & 40.79 & 178.32 \\
    1     & 35.14 & 12.04 & 40.90 & 173.84 \\
    \bottomrule
    \end{tabular}
\end{table}

\subsection{Hyperparameter Exploration}
We study three modules: single, pairwise, and multi-preference optimization. As single preference is well-studied, we set its weights to $\lambda = 1.0$ (response) and $\alpha = 0.5$ (visual/context). We then vary $\gamma$ to control pairwise and multi-preference integration. As shown in \autoref{tab:parameters}, MCM-DPO performs best on three of four metrics when $\gamma = 0.2$, which we adopt in \autoref{eq:target}. Thus, we adopt $\lambda = 1.0$, $\alpha = 0.5$, and $\gamma = 0.2$ in \autoref{eq:target}.

\subsection{Representation Visualization}

\begin{figure}[!ht]
  \centering
  \begin{subfigure}{0.25\linewidth}
    \includegraphics[width=1\textwidth]{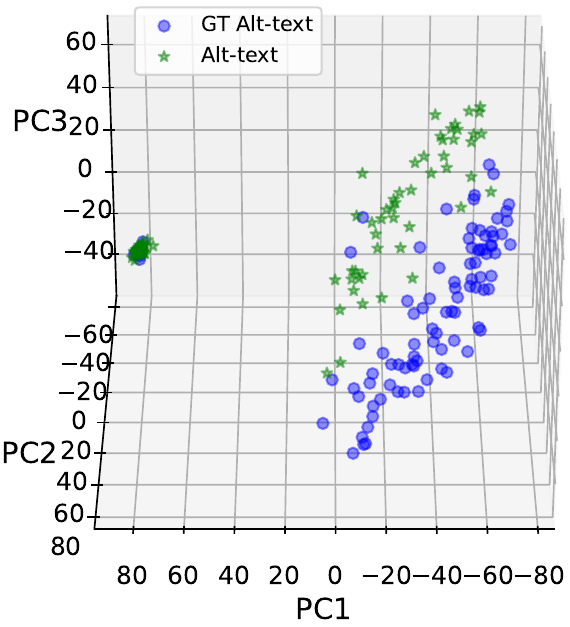}
    \caption{LLaVA}
  \end{subfigure}
  \hspace{-6pt}
  \begin{subfigure}{0.25\linewidth}
    \includegraphics[width=1\textwidth]{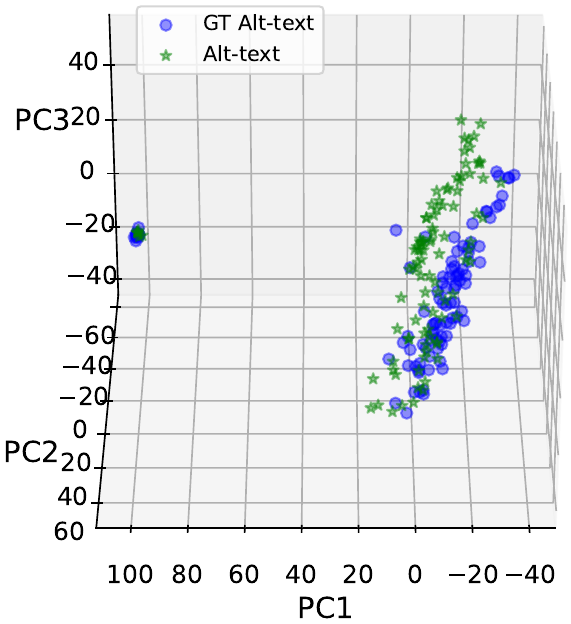}
    \caption{ + SFT}
  \end{subfigure} 
  \hspace{-6pt}
  \begin{subfigure}{0.25\linewidth}
    \includegraphics[width=1\textwidth]{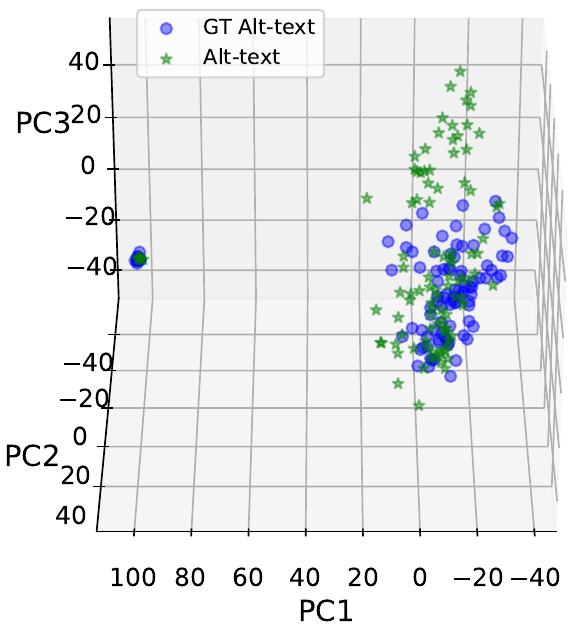}
    \caption{ + DPO}
  \end{subfigure}
  \hspace{-6pt}
  \begin{subfigure}{0.25\linewidth}
    \includegraphics[width=1\textwidth]{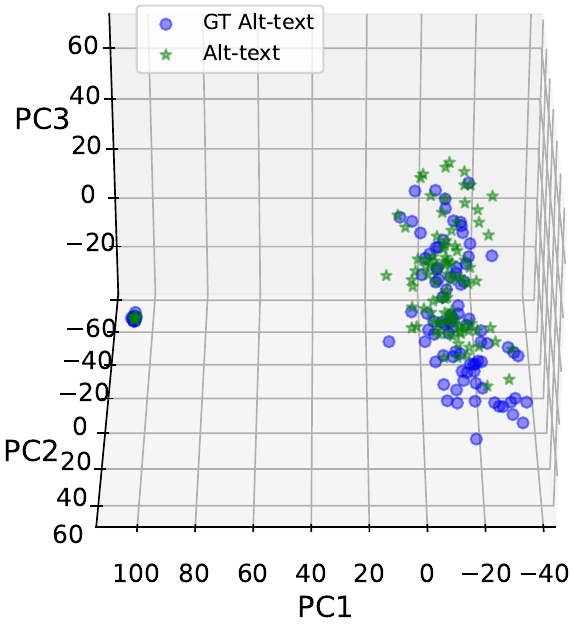}
    \caption{ + MCM-DPO}
  \end{subfigure} 
  \caption{Representation visualization. ``GT Alt-text'' and ``Alt-text'' refer to the ground-truth and model-generated alt-text, respectively.
}
\label{fig:perfm_visaul}
\end{figure}

To intuitively show MCM-DPO’s effectiveness, we visualized alt-text representations from MCM-DPO, DPO, SFT, and LLaVA, alongside ground-truth. We visualized representations of the alt-text generated by these models and the ground-truth alt-text.
Ideally, high-quality alt-text should closely match ground-truth semantics, yielding similar representations. We randomly sampled 100 PAlt-test examples, encoded both generated and ground-truth alt-text using LLaVA (Vicuna’s final token embedding), and applied PCA for visualization, as shown in \autoref{fig:perfm_visaul}.

\textbf{Observations}: 
LLaVA’s alt-text initially differs significantly from ground-truth, but this gap narrows with SFT, improves with DPO, and is minimized by MCM-DPO. It shows that the MCM-DPO best aligns outputs with ground-truth.

\section{Conclusion}
In this work, we focus on the task of alt-text generation, aimed at enabling visually impaired users to understand image content through specialized text-reading tools. Specifically, we address the challenges posed by inconsistent annotation standards and high noise in alt-text created by social media users for their own images. To tackle these issues, we construct preference pairs based on three key aspects: alt-text, post-text (context), and images, and propose a novel Multi-faceted Cross-modal Direct Preference Optimization (MCM-DPO) model.
MCM-DPO optimizes preferences across Single, Pairwise, and Multi-Preference dimensions. Given the scarcity of high-quality datasets, we developed two large-scale datasets from Twitter and Pinterest. Experimental results demonstrate that MCM-DPO consistently outperforms existing DPO and SFT methods.

\begin{acks}
This research/project is supported by the National Research Foundation, Singapore under its National Large Language Models Funding Initiative (AISG Award No: AISG-NMLP-2024-002). Any opinions, findings and conclusions or recommendations expressed in this material are those of the author(s) and do not reflect the views of National Research Foundation, Singapore.
This work is also supported by the National Natural Science Foundation of China (No. U24B20181) and Shanghai Pilot Program for Basic Research - Fudan University 21TQ1400100 (22TQ018). 
\end{acks}

\bibliographystyle{ACM-Reference-Format}
\bibliography{sample-base}

\appendix

\section{Preference Optimization for Hallucination Mitigation}
To evaluate whether our proposed MCM-DPO framework can be effective on other public benchmark datasets, we adapted MCM-DPO for multimodal hallucination mitigation tasks where the input does not contain post-text. The corresponding preference optimization tasks for this setup are visual question answering (VQA) and image captioning.

\textbf{Training Datasets}
We utilize the RLHF-V-Dataset released by \citet{yu2024rlhf} along with 5k training samples as our training dataset. This dataset includes tasks for both VQA and image captioning.

\textbf{Benchmarks and Evaluation Metrics}
We conducted evaluations on the following three popular multimodal hallucination datasets: 

\begin{itemize*}
    \item \texttt{Object HalBench (ObjHal)}~\citep{rohrbach2018object} is a commonly used benchmark for assessing object hallucination. 
    Metrics: In line with ~\cite{yu2024rlhf,wang2024mdpo}, we report two metrics: the response-level hallucination rate (\texttt{R.}) and the mention-level hallucination rate (\texttt{M.}).

    \item \texttt{MMHal-Bench (MMHal)}~\citep{sun2023aligning}  is a question-answering benchmark encompassing 8 question categories and 12 object topics. It evaluates response quality (\texttt{O.}) and hallucination rates (\texttt{R.}) using GPT-4 as the assessment model.

    \item \texttt{AMBER}~\citep{wang2023llm} was designed to be evaluated without LLM assistance. 
    Following previous works~\citep{wang2024mdpo}, we only consider the generative tasks.
     \textit{Metrics}:
    (a) CHAIR~\citep{rohrbach2018object} (\texttt{CHAIR}), a widely used metric for evaluating hallucinations.
    (b) Object coverage of responses (\texttt{Cover}), measuring the proportion of objects mentioned in the response.
     (c) Response-level hallucination (\texttt{Hal}), denotes the proportion of responses with hallucinations.
    (d) Human cognition hallucination (\texttt{Cog}). measuring the likelihood of VLMs that generate the objects from predefined hallucinatory target objects of human cognition.

\end{itemize*}

\textbf{Results}
Similar to the alt-text task, we chose LLava-1.6 as our backbone model and conducted preference optimization training using DPO and MCM-DPO. The experimental results are presented in \autoref{tab:hallu}.
MCM-DPO demonstrates its effectiveness in significantly reducing hallucination rates while preserving text generation quality. On the AMBER dataset, MCM-DPO achieves a hallucination rate of 27.9\%, representing a relative reduction of 42.6\%, while maintaining or slightly enhancing object coverage. Similar trends are observed on the MMHal dataset. Furthermore, on the ObjHal dataset, MCM-DPO achieves the lowest hallucination rates at both the response and mention levels.
This indicates that the proposed MCM-DPO not only works effectively on the alt-text task but also performs well on some downstream tasks with publicly available datasets.

\begin{table}[htbp]
  \centering \small
  \setlength{\tabcolsep}{2pt}
  \caption{Results of MCM-DPO on multimodal hallucination. Bold is the best performance. `↓' indicates that lower values are better, while `↑' denotes that higher values are better.}
  \vspace{-8pt}
    \begin{tabular}{lcccccccc}
    \toprule
    \multirow{2}[2]{*}{Model} & \multicolumn{2}{c}{ObjHal} & \multicolumn{2}{c}{MMHal} & \multicolumn{4}{c}{AMBER} \\
    \cmidrule(lr){2-3}\cmidrule(lr){4-5}\cmidrule(lr){6-9}
          & R.↓ & M.↓ & O.↑ & R.↓ & CHAIR↓ & Cover↑ & Hal↓  & Cog↓ \\
    \midrule
    GPT-4V~\cite{gpt4V} & 13.6 & 7.3 & - & 31.3 & 4.6 & 67.1 & 30.7 & 2.6 \\
    \midrule
    LLaVA-1.6  & 14.1  & 7.4   & 2.8   & 42.7  & 8.3   & 61.0 & 48.6  & 4.2 \\
    +DPO  & 11.0  & 6.6   & 2.7   & 43.8  & 5.9   & 61.0 & 38.9  & 3.0 \\
    +MCM-DPO & \textbf{8.9}      & \textbf{5.2 }    & \textbf{2.9}       & \textbf{38.5}     & \textbf{4.7}     & \textbf{61.2}     & \textbf{27.9}     & \textbf{2.3} \\
    \bottomrule
    \end{tabular}
  \label{tab:hallu}
\end{table}

\section{Prompt for Data Construction}

The grammar correction prompt is shown below:
\begin{mdframed}[linewidth=0.5pt, roundcorner=3pt]
\raggedright 
\footnotesize 
 Your task is to check the given text for grammatical errors. Are there any grammatical errors in the following text? \\
\{alt-text\} \\ 
Please generate the final judgment strictly in the following format: \\
(1) If there are any grammatical errors, then output `True'; otherwise output `False'. \\
(2) If your answer is True, then you need to provide the text after 
correcting the grammar issue in the following format: ``Corrected text:'' \\
Make as few modifications as possible. \\
If your answer is False, then no further changes are needed. \\
\end{mdframed}

\paragraph{The prompt for the rejected alt-text generation is shown below:}

\begin{mdframed}[linewidth=0.5pt, roundcorner=3pt]
\raggedright 
\footnotesize 

Given the best alt-text of an image and context. If you are writing the alt-text for the given image, what kind of suboptimal alt-text will you write? \\
Context:  \\
\{context\} \\
Best alt-text: \\
\{alt\_text\} \\
Suboptimal alt-text: \\
\end{mdframed}

\section{Implementation Details}
We selected LLaVA-1.6~\cite{liu2023improved} as the base model. LLaVA-1.6 utilizes CLIP's ViT~\cite{radford2021learning} as the vision encoder and the 7B version of Vicuna~\cite{liu2023improved} as the LLM backbone. We fine-tuned the LLaVA-1.6 (7B) model using the mixing of large-scale supervised fine-tuning datasets PAlt and TAlt, which consist of 202K samples collected from the Twitter and Pinterest platforms, respectively.
Supervised fine-tuning on such large-scale datasets helps the base model acquire domain-specific and task-specific knowledge, facilitating efficient preference optimization.
The training was conducted for one epoch, using an image resolution of 336, a learning rate of 2e-5, and a batch size of 128.

\noindent
\textit{Preference optimization }
We applied DPO and MCM-DPO to the SFT model, using the TAlT and PAlt preference dsatasets for alignment training, respectively. The training ran for three epochs with a learning rate of 5e-7 and a batch size of 64.

\noindent
\textit{License }
The data, code, and models are for research use only and must adhere to the LLaMA and Vicuna license terms. The dataset is licensed under CC BY-NC 4.0, allowing only non-commercial use, and any models trained with it should strictly be used for research purposes.

\section{Cost‑Effectiveness Analysis}

\textbf{Resource Overhead:}
Compared with standard DPO, which uses a single loss function, MCMDPO employs seven. Although this means six additional loss terms, all seven share the same forward and backward pass, so no extra model components or inference paths are introduced.

\textbf{Training Time:}
On the same dataset and settings (4 $×$A100 80GB, batch size$=64$, 3 epochs), standard DPO finishes training in 4.6 hours, whereas MCMDPO takes 5 hours—an increase of 8.7\%. Inference latency remains unchanged. The performance gains, however, are substantial: on the Pinterest dataset, ROUGE‑L rises from $32.71$ to $39.54$, and CIDEr jumps from $157.73$ to $207.98$. Given these improvements, the extra training time is acceptable.

\section{Case Study}
\begin{figure}
    \centering
    \includegraphics[width=0.98\linewidth]{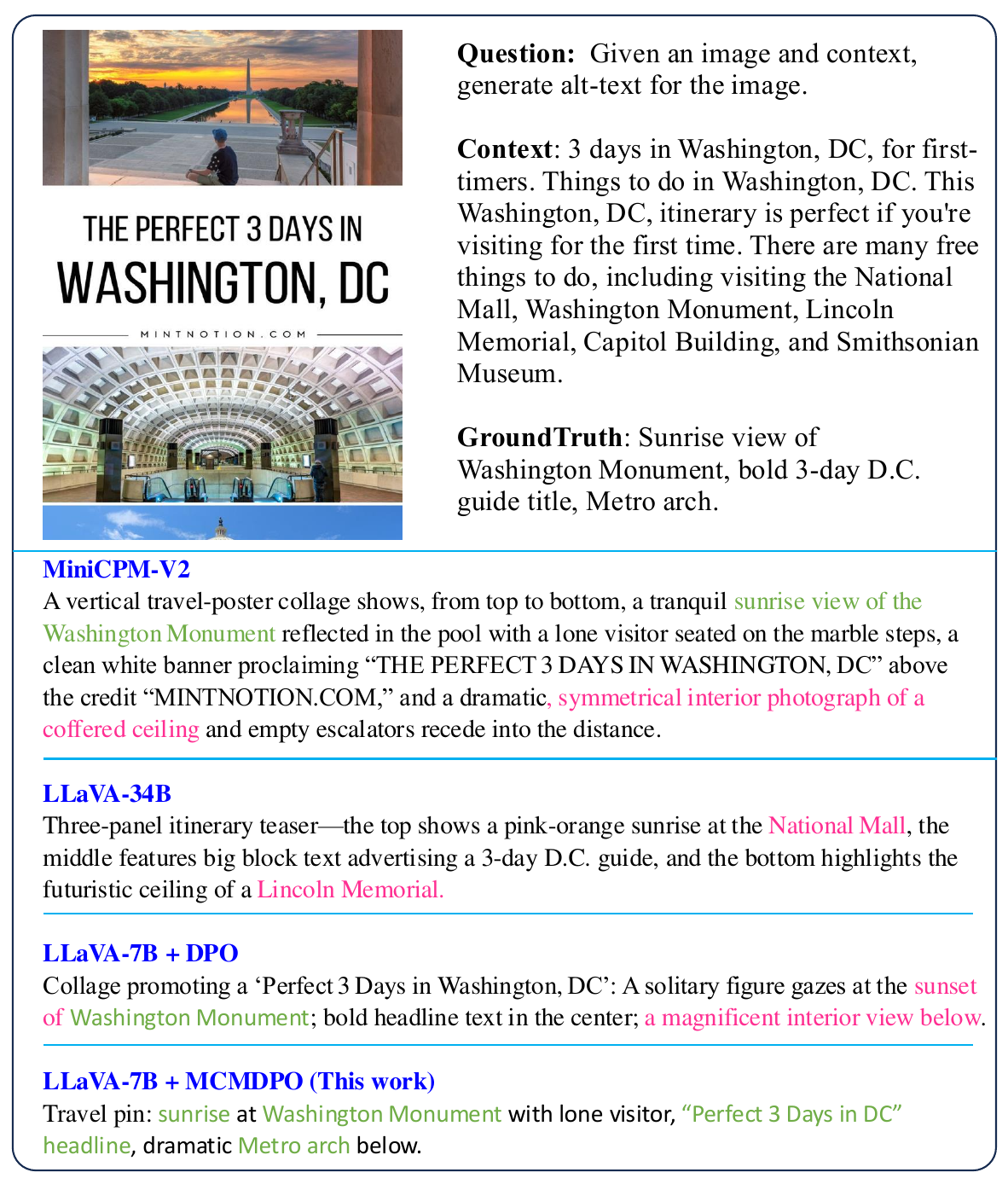}
    \vspace{-8pt}
    \caption{Comparison of alt-text generated by different models given the same image and Twitter-text context. Pink text indicates hallucinations or errors, while green text highlights correct content.
    }
    \label{fig:case1}
\end{figure}

To better demonstrate the advantages of our MCMDPO approach over models that are either untrained or fail to consider multifaceted preference optimization, we present comparative results across real-world examples using different models. These include untrained models (e.g., MiniCPM-V-2 and LLaVA-34B) and standard DPO-based methods (e.g., LLaVA+DPO), as shown in \autoref{fig:case1}.

\noindent
Our \textbf{observations} are as follows:

(1) Models without alt-text training, such as MiniCPM-V-2 and LLaVA-34B, tend to produce overly detailed descriptions of the image, which deviates from the concise and context-aware nature that the alt-text task typically requires.

(2) LLaVA+DPO, a key baseline in our MCM framework, shows a tendency to “overlook the image” and instead generate responses primarily based on the given textual context. This behavior likely stems from DPO’s focus on optimizing response-level preferences alone, leading to an over-reliance on textual inputs.

(3) LLaVA+MCMDPO achieve best performance. Our method employs Multifaceted Cross-Modal Preference Optimization, which includes:
(a) Single-aspect preference modeling, optimizing preferences individually over the image, response, or context;
(b) Pairwise preference modeling, capturing preferences between any two aspects among image, response, and context;
(c) Multi-aspect preference modeling, jointly optimizing preferences across all three aspects simultaneously.
By introducing hierarchical preference signals across both visual and textual modalities, our approach enables the model to more effectively compare and align information across image, response, and context. This leads to more precise preference learning and ultimately produces responses that are better grounded in all relevant aspects.

\end{document}